\documentclass[sigconf]{aamas}

\usepackage{balance} %
\usepackage[nolist]{acronym}
\usepackage{tikz}
\usetikzlibrary{positioning}

\begin{acronym}
\acro{AI}{Artificial Intelligence}
\acro{XAI}{Explainable AI}
\acro{HLEGAI}{European Commission’s High Level Expert Group on Artificial Intelligence}
\acro{EASA}{European Union Aviation Safety Agency}
\acro{DL}{Deep Learning}
\end{acronym}

\setcopyright{ifaamas}
\acmConference[AAMAS '23]{Proc.\@ of the 22nd International Conference
on Autonomous Agents and Multiagent Systems (AAMAS 2023)}{May 29 -- June 2, 2023}
{London, United Kingdom}{A.~Ricci, W.~Yeoh, N.~Agmon, B.~An (eds.)}
\copyrightyear{2023}
\acmYear{2023}
\acmDOI{}
\acmPrice{}
\acmISBN{}

\acmSubmissionID{???}

\title[Inherently Interpretable Knowledge Representation for Trustworthy Human-Machine Teaming]{Flexible and Inherently Comprehensible Knowledge Representation for Data-Efficient Learning and Trustworthy Human-Machine Teaming in Manufacturing Environments}

\author{Vedran Galeti{\'c}}
\affiliation{
  \institution{Airbus Central R\&T AI Research Team}
  \city{Filton}
  \country{United Kingdom}}
\email{vedran.galetic@airbus.com}

\author{Alistair Nottle}
\affiliation{
  \institution{Airbus Central R\&T AI Research Team}
  \city{Filton}
  \country{United Kingdom}}
\email{alistair.nottle@airbus.com}

\begin{abstract}
Trustworthiness of artificially intelligent agents is vital for the acceptance of human-machine teaming in industrial manufacturing environments. Predictable behaviours and explainable (and understandable) rationale allow humans collaborating with (and building) these agents to understand their motivations and therefore validate decisions that are made. To that aim, we make use of Gärdenfors’s cognitively inspired Conceptual Space framework to represent the agent’s knowledge using concepts as convex regions in a space spanned by inherently comprehensible quality dimensions. A simple typicality quantification model is built on top of it to determine fuzzy category membership and classify instances interpretably. We apply it on a use case from the manufacturing domain, using objects’ physical properties obtained from cobots’ onboard sensors and utilisation properties from crowdsourced commonsense knowledge available at public knowledge bases. Such flexible knowledge representation based on property decomposition allows for data-efficient representation learning of typically highly specialist or specific manufacturing artefacts. In such a setting, traditional data-driven (e.g., computer vision-based) classification approaches would struggle due to training data scarcity. This allows for comprehensibility of an AI agent’s acquired knowledge by the human collaborator thus contributing to trustworthiness. We situate our approach within an existing explainability framework specifying explanation desiderata. We provide arguments for our system’s applicability and appropriateness for different roles of human agents collaborating with the AI system throughout its design, validation, and operation.
\end{abstract}

\keywords{Knowledge Representation and Reasoning, Conceptual Spaces, Explainable AI, Trustworthy AI, Comprehensibility, Interpretability, Human-Machine Teaming, Embodied Intelligent Agents, Frugal AI} %

\newcommand{\BibTeX}{\rm B\kern-.05em{\sc i\kern-.025em b}\kern-.08em\TeX}

\begin{document}

\pagestyle{fancy}
\fancyhead{}

\maketitle

\section{Introduction}
Use of \ac{AI} is increasing across industry, often providing performance approaching, or even surpassing, that of humans on cognitive tasks for specific applications (e.g., \cite{szegedy2015going,komorowski2018artificial}). This in turn leads to an increase in the use of embodied \ac{AI} agents cooperating with humans and becoming embedded on critical systems. There is an increasing need to build trust within these teams. For instance, within the frame of aerospace manufacturing and operations, \ac{AI} can assist workers in practical tasks such as: optimising scheduling of resources; automated visual inspections; and natural language interactions for receiving commands and reporting events; as well reducing user's cognitive load and supporting automated decision making in fleet management.

Whilst performance gains can be clear and measurable, ensuring acceptance of these technologies, and thereby realising those gains, is difficult. Explainability is one of key elements in building appropriate trust (i.e., not complacency), which is necessary for their acceptance.

Current high-performing \ac{AI} systems, often based on deep neural network models (i.e., \ac{DL}), tend to be \lq black boxes\rq, shrouding their internal knowledge and decision making processes, which in turn may not be readily accessible or interpretable to human users and subjects interacting with, or developing, them. \ac{XAI} is widely seen as one of the cornerstones of \ac{AI} trustworthiness. Indeed, the \ac{HLEGAI} specifies \lq explainability\rq{} (or \textit{explicability}) as one of their four ethical principles of fundamental trustworthiness of \ac{AI}, with the others being: respect for human autonomy, prevention of harm, and fairness~\cite{ecai} (and \ac{XAI} can play a part in ensuring those other three principles). Furthermore, the Group specifies seven key requirements that are to be addressed throughout an \ac{AI} product’s life-cycle, from which ‘transparency’, ‘accountability’, and ‘human agency and oversight’ are ones clearly related to explainability. These guidelines are upheld by the \ac{EASA}, focusing on challenges around certification that black-box \ac{AI} models impose, echoing explainability as one of the three main components of trustworthy \ac{AI}~\cite{easa_ml}, and fully recognising human-centricity in its \ac{AI} Roadmap~\cite{easa_roadmap}.

While the interest in XAI research has been tremendous both in academia and industry (e.g., \cite{lipton2018mythos,doshi2017towards,arrieta2020explainable,gunning2019darpa}), the outputs have primarily been focussed on \textit{post-hoc} explainability methods and techniques, i.e., providing explanations of an already developed high-performing yet \lq explanatorily opaque\rq{} machine learning systems. Instead, we attempt inherent (or intrinsic) explainability by incorporating interpretability requirements early in the system development cycle and focusing on building inherently interpretable~\footnote{As a note, we do not equate inherent interpretability with algorithmic or model transparency characteristic~\cite{lipton2018mythos}. We adopt a more general definition of interpretability as \lq \textit{the ability to explain or to present in understandable terms to a human}\rq{}~\cite{doshi2017towards}.} and understandable \ac{AI} systems. In this paper we demonstrate this through a more inherently explainable knowledge representation of the \ac{AI} agent.

We model the agent's knowledge using a complementary combination of abstracted information from the agent's sensors and openly available commonsense general knowledge sources, subject to simple heuristic engineering. By representing both classes of knowledge on the same representation framework, typical for \textit{hybrid AI} approaches, allows for knowledge modelling flexibility across application domains. Also, it addresses challenges of (high-performant) computer vision-based approaches of object classification pertaining to data scarcity for rare and specialist objects (not uncommon in aerospace manufacturing) and account of its rationale and outputs in a human-understandable actionable way.

\section{Interpretable Knowledge Representation}
\label{XKR}

One means of increasing trustworthiness, and consequent certifiability\footnote{By \textit{Certifiability} we mean the ability for a system to be certified for use in a regulated environment.}, of an AI agent teaming with humans in operational environments is through inherently interpretable knowledge representation modelling. This allows for the agent's comprehensibility, i.e., the ability to represent its acquired knowledge in human-understandable terms~\cite{arrieta2020explainable}.

Symbolic representations are a classical way to represent knowledge due to symbols' intrinsic meaning. Although symbols are amenable for computational approaches involving logical calculus, one challenge of this level of representation is modelling the intelligent agent's concept acquisition~\footnote{It is, however, fair to acknowledge some promising steps forward in modelling interpretable concept acquisition and representation on the neural implementation level, e.g., see \cite{blazek2021explainable}.}. Artificially intelligent agents, especially those that are embedded in complex environments and perform higher cognitive tasks, are predominantly based on deep neural networks, excelling at learning from statistical regularities of sensory inputs. Human mind employs prominent qualities of both of these representations.

Therefore, another challenge is mapping between the continuous space of knowledge representation engendered by learnt network models and the symbolic representations. Systematic neuro-symbolic mapping, or an intermediary knowledge representation interfacing with both levels, would alleviate one of the major challenges of neural representations, namely, their opacity and consequent incomprehensibility of learnt knowledge, standing in the way of the AI's trustworthiness and certifiability.

\subsection{Conceptual Knowledge Representation}
\label{CS}
The semantic theory of Conceptual Spaces~\cite{gardenfors2004conceptual,gardenfors2014geometry} is an apt formalism in bridging the neuro-symbolic gap in knowledge modelling, proved useful in various application domains~\cite{zenker2015applications}. Concepts have inherent meaning, often denoted by corresponding symbols (those concepts that are languageable), whilst retaining their continuous nature arising from neural sensing of environmental input. The framework represents concepts as convex regions in a geometric space spanned by quality domains. A concept (e.g., \textit{apple}) is described by pertaining ranges across quality domains (e.g., \textit{colour}, \textit{size}, \textit{taste}) and encompasses instances (e.g., \textit{this green and sour apple sitting on my desk}) represented as vectors characterised by specific properties (e.g., \textit{green}, \textit{sour}).

The Conceptual Spaces framework adopts basic tenets of cognitive semantics (inherited in turn from cognitive psychology), thus assuming prototypical effects in categorisation~\cite{rosch1978principlesInbook,rosch1976basic,rosch1975family} and schematicity~\cite{lakoff2008women}. Conceptual algebraic operations and space mappings can be used to describe metaphoric and metonymic communication operations, while modelling concept combinations and quantifying similarity\footnote{Conceptual Spaces also allow for discerning between similarity (e.g., \textit{car} and \textit{van}) and relatedness (e.g., \textit{car} and \textit{driver}), which is one of the challenges of distributional semantics approach.} arise naturally from the very structure of the geometric space.

Moreover, it has been empirically demonstrated that neural populations exhibit geometric (hexadirectional) organisation of conceptual knowledge in the mind, demonstrated on spatial tasks in rodents \cite{hafting2005microstructure} as well as humans on spatial and, crucially, abstract non-spatial tasks \cite{constantinescu2016organizing}. It is suggested that cognitive maps of concepts may utilise place cells for concept indexing within a geometric space spanned by the hippocampal-entorhinal grid cell system. That makes Conceptual Spaces a valuable candidate for a neuroscientifically (along with cognitively) motivated knowledge representation framework~\cite{bellmund2018navigating}.

\subsection{Crowdsourced Knowledge Base}
\label{CN}
Humans are able to rapidly construct hyper-hypotheses based on remarkably little data~\cite{tenenbaum2006theory,tenenbaum2011grow} and utilise them as background knowledge and constraints in various tasks such as ad-hoc classification. Capturing the structure and content of commonsense knowledge is a challenging task in modelling an artificially intelligent agent. As a remedy, manually-crafted and crowdsourced knowledge bases can be used as a proxy. They are typically knowledge graphs such as ontologies (e.g., \cite{lenat1989building}) or hierarchically organised synsets~\cite{miller1995wordnet}.

ConceptNet~\cite{speer2017conceptnet} is an openly-available knowledge graph constructed through orchestrated input from both structured manually-crafted sources (e.g., Wiktionary, OpenCyc~\cite{lenat1989building}) and crowdsourcing campaigns. Despite heterogeneous sources and a vast number of concepts (8 million nodes) and relations between them (21 million edges), the knowledge base is remarkably tidy and manageable with only 36 consolidated relation types. Some of these relations such as \textit{UsedFor}, \textit{MadeOf}, and \textit{PartOf} are of particular relevance for our use case in the aerospace manufacturing domain.

As we will see in the upcoming sections, utilisation properties that are acquired from this knowledge base shall be instrumental in modelling concepts as utilisation properties tend to be more relevant for classifying manufacturing objects than their surface physical properties.

\subsection{Typicality Model}
\label{muw}
Whilst some very developed and faithful Conceptual Space implementations exist~\cite{bechberger2017thorough}, we draw inspiration from the Conceptual Space framework and property decomposition-based representation of concepts that it employs, and propose utilising a simple model for interpretable fuzzy~\cite{zadeh1996fuzzy} classification. The model, called \textit{\textmu w-model}~\cite{galetic2011aggressive}, relies on the prototype theory~\cite{rosch1978principlesInbook} and empirical theories on causal status~\cite{ahn2000acausal,ahn2000bcausal} and concept centrality~\cite{sloman1998feature}.

\begin{figure}
\includegraphics[width=\columnwidth]{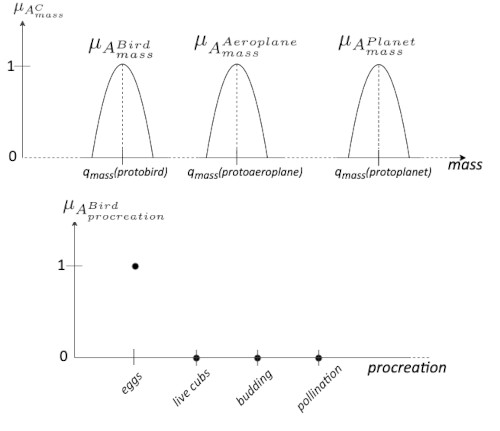}
\caption{Illustration of the membership functions $\mu$ for a continuous quality dimension $mass$ across three example concepts (prototypical instances are assumed to bear prototypical value of the property) and of a nominal quality dimension $procreation$ for the $Bird$ concept.}
\centering
\label{mus}
\end{figure}

\begin{figure}
\includegraphics[width=\columnwidth]{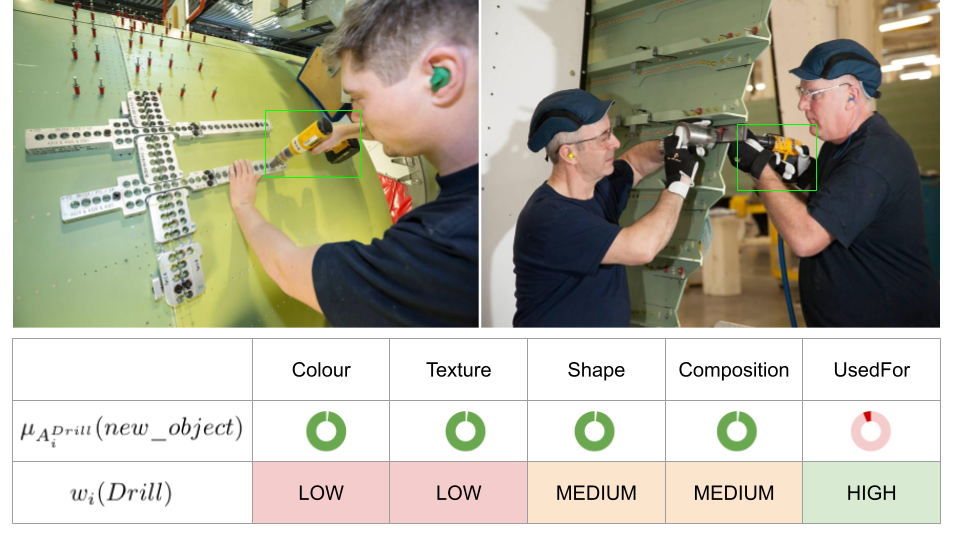}
\caption{The artificial agent wants to classify a new object (upper right image) and extracts its physical and utilisation properties. The table (below) illustrates the $\mu$ and $w$ values for the ‘Drill' concept. Although the new object is physically similar to the learnt ‘Drill' prototype (upper left image shows a typical instance of ‘Drill'), it will not be classified as one due to incompatibility of the utilisation property bearing the highest weight.}
\label{drill_riveter}
\centering
\end{figure}

The model's name originates from two parameters that are used to describe an instance's typicality status in the frame of a concept, namely: the membership function \textit{\textmu} quantifying typicality for one quality dimension (see Fig.~\ref{mus}); and the weight \textit{w} per quality dimension. It is thereby possible to ascertain and explain that, for example, although an object observed in the manufacturing environment has surface similarities to the prototype of a previously learnt class, the facts that it is used for a different activity (see \S~\ref{CN}) and that the utilisation property is much weightier than surface properties in the manufacturing environment jointly steer its classification to another class (e.g., Fig.~\ref{drill_riveter}).

These parameters are learnt in different ways, yet both are based on the agent's property detection capabilities (e.g., visual sensor) accompanied by tutor-provided labels of encountered instances. Based on instances' labels, and quality dimension value distributions, the agent is able to quantify \textit{\textmu} frequentistically (e.g., assuming the prototype has the most occurring value), while \textit{w} is quantified by using an empirically confirmed~\cite{galetic2016phd} hypothesis, based on concept centrality, on inverse relationship between the dimension's weight on one side and its variability of instances' property values across the category on the other (e.g., \textit{colour} tends to be more important for natural kinds~\cite{rosch1976basic} such as \textit{apple} than for artefacts such as \textit{car} as the latter can usually be in an arbitrary colour).

Each instance is represented as a vector as shown in Eq.~\ref{eqr}:

\begin{equation}
\label{eqr}
    \overrightarrow{r_C}(c)=\sum_{i}\sqrt{\frac{w_i(C)}{\sum_{j}w_j(C)}}\mu_{A^C_i}(q)\cdot \overrightarrow{e_i}
\end{equation}

where $i$ is a quality dimension (e.g., \textit{width}), $\overrightarrow{e_i}$ are basis vectors spanning the space, $j$ is a quality domain (e.g., \textit{size}), $q$ is an instance (e.g., \textit{this apple}), $w_i(C)$ is the weight of the quality dimension $i$ for the concept $C$ (e.g., \textit{apple}), and $\mu_{A^C_i}(q)$ is the typicality measure of the instance $q$ for the concept $C$ with respect to the quality dimension $i$ (e.g., the representativeness of  ‘\textit{this zebra-striped ball}’ for the concept \textit{zebra} with respect to the quality dimension \textit{texture}; which, as a note, would be high for this dimension, but extremely low for virtually any other dimension).

Typicality of the instance \textit{c} with respect to concept \textit{C} is the second norm of vector in Eq.~\ref{eqr}, thus is calculated as in Eq.~\ref{eqR}:

\begin{equation}
\label{eqR}
    R_C(c)=\|\overrightarrow{r_C}(c)\|=\sqrt{\sum_{i}\frac{w_i(C)}{\sum_{j}w_j(C)}\mu^2_{A^C_i}(q)}
\end{equation}

while it is sensible to treat the instance as one pertaining to that concept for which its typicality is the highest, as in Eq.~\ref{eqCat}.

\begin{equation}
\label{eqCat}
    Cat(c)=\max_{C}R_C(c)
\end{equation}

As the system is able to flag any disputable classifications (e.g., those for which the difference between the maximal and the second largest typicality from Eq.~\ref{eqR} is not above a threshold) in an interpretable way, it is possible to investigate the case using intrinsically interpretable explanation of the rationale as, e.g., a developer debugging the system or an operator on the shop floor.

\section{Manufacturing Use Case}
To examine the feasibility of our model in a \lq real-world\rq{} scenario, we applied the described knowledge modelling approach to an aerospace manufacturing process. There is the potential to make improvements to existing capability as well as develop new ones though the use of collaborative robots, or \lq cobots\rq{}. These \lq cobots\rq{} will improve quality and free up skilled workers to focus on their specialities. Therefore, explainability of decisions from robotic assistants is vital to building acceptance and trust and enabling effective collaboration, allowing these benefits to be realised.

In order to build within cobots an awareness of their environment they need to be able to rapidly and efficiently recognise objects which they may not have seen previously as well as with clear and understandable explanations for this recognition. By modelling the artificial agent's knowledge interpretably, using the Conceptual Space framework \cite{gardenfors2004conceptual}, we reduce the need for post-hoc explanation.

As an example, a cobot may classify an object as a \textit{drill} because it because it is of similar size, shape, colour, and material composition, as examples used during the training process. Capturing utilisation properties, which can indicate for instance that the item is used \textit{for drilling}, allows us to produce classification which is by its very nature interpretable. For example, the operator may obtain a natural language output such as ‘\textit{I believe this is a drill as it looks similar to other drills I’ve seen in the past, and it is used for drilling}’, accompanied by a visualisation with associated typicality measures. Moreover, if an object bears surface similarities to previously seen instances of a \textit{drill}, but is perceived to be used \textit{for riveting} would instead be classified as a riveter. This reflects that utilisation properties of industrial artefacts are more heavily weighted in a classification than the surface properties (Fig.~\ref{drill_riveter}).

\subsection{Setup}
Industrial environments can be extremely sensitive to disruption, particularly with just-in-time manufacturing processes. Therefore, current industrial setups do not readily allow for the deployment of experimental robotics on the factory floor. To combat this we have instead used a simulated environment of a cobot in a factory setting to provide training data, as well as to get feedback and validation of the proposed methodologies. We have used the \textit{Webots Open Source Robotics Simulator} \cite{webots} to create, in the first instance, a simple environment with a controllable \textit{e-puck} robot \cite{epuck} equipped with a simple sensor package, such as a standard vision sensor (i.e., a camera) as well as a time-of-flight sensor.

The simulation environment is populated with a variety of objects, using simplistic idealised examples such as ‘Green Ball’ or ‘Red Cube’, to allow for easier recognition, but still \lq relatable\rq{} and inherently interpretable objects (see Fig. ~\ref{demo}). As we enhanced the \ac{AI} system, we introduced additional objects to represent real-world, ecologically relevant, objects such as a hammer or screwdriver.

To further augment the properties available, we attach custom properties to the objects within the simulator (e.g., stripy texture). This assists in the determining of ground truth values for colour, texture, utilisation properties, etc. These values can be extracted from the simulated environment programmatically along with the labels of the objects to provide a training set from which to learn a conceptual representation using inherently explainable and interpretable terms.

A conceptual space is learned by Voronoi tessellation~\cite{gardenfors2001reasoning} around prototypes. In the simpler simulation environment, the objects are considered to already be idealised prototypes in order to make it easier to carry out proof-of-concept space construction. With the migration to a more complex simulation environment, prototypes are calculated as centroids of the labelled instances used during training.

The robot can then be manipulated by a user to move through the simulation, and the field-of-view of the camera is determined. Any objects coming within this field of vision will then trigger the \textit{Webots} software to forward the image to our API, alongside the ground truth properties relating to that object.

In a real-world implementation, capturing of images and detection of the objects therein would be carried out by state-of-the-art object / region of interest detection algorithms. As these are not the focus of our research, we have instead extracted ground truth data and imagery from the simulation environment itself. This modular approach allows for algorithms (such as property detectors) to be inserted as needed, and users can select the most appropriate algorithm for the use case being exploited.

This geometric representation level can thus be treated as an intermediary "middle ground" between the symbolic and neural levels by, on the one hand, abstracting sensory inputs onto appropriate points in a conceptual space, whilst, on the other hand, grounding symbols onto concept prototypes (as suggested in, e.g., \cite{tabakowska2005gramatika}).

\subsection{Property Extraction for Interpretable Classification}
\label{XC}

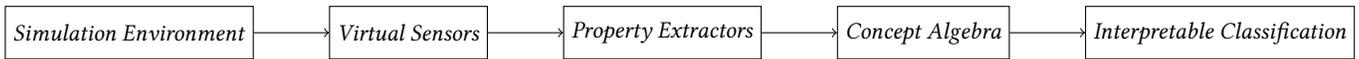
\begin{figure*}
\begin{tikzpicture}[
        squarenode/.style={rectangle, draw=black, minimum size=7mm, align=left}]
    
    \node[squarenode] (SimEnv) {\textit{Simulation Environment}}; 
            \node[squarenode] (virtSens) [right=of SimEnv] {\textit{Virtual Sensors}}; 
            \node[squarenode] (propExtract) [right=of virtSens] {\textit{Property Extractors}};
            
            \node[squarenode] (conAlg) [right=of propExtract] {\textit{Concept Algebra}}; 
            \node[squarenode] (interpClass) [right=of conAlg] {\textit{Interpretable Classification}};
            \draw[->] (SimEnv.east) -- (virtSens.west);
            \draw[->] (virtSens.east) -- (propExtract.west);
            \draw[->] (propExtract.east) -- (conAlg.west);
            \draw[->] (conAlg.east) -- (interpClass.west);
        \end{tikzpicture}
        \caption{Data flow in the manufacturing use case. Currently, simulated environment is used in the proof-of-concept scenario. Virtual, visual-based sensors are used to capture objects in the environment. Captured images are fed into the property extractor modules (e.g., for \textit{colour}, \textit{texture}, etc.), each dedicated to extracting the value corresponding to the pertaining quality dimension (e.g., \textit{hue}, \textit{stripy}, etc.). It is worth mentioning that some quality domains, whose extraction is more challenging due to limitations of the simulated environment (e.g., \textit{size}), are quantified using the simulated environment specification directly, which is a temporary fix in line with the fact the focus of our work is not as much on the property detectors and computer vision as on concept building and class inference proof of concept in the "back end". An instance described by its (standardised) property values is represented as a vector and its representativeness measures for candidate classes quantified using the \textit{\textmu w-model}.}
        \label{dataFlow}
        
\end{figure*}

Separate property detectors are applied to extract properties of the observed object. These take two broad categories of detectors: physical property detectors and utilisation property detectors. Basic physical properties can be inferred from the sensors on the robotic platform. Examples of properties we have experimented with include: texture (using \textit{Concept Activation Vectors}~\cite{kim2018interpretability} to determine distinctive textural properties of an object, e.g., stripey, smooth, etc.); colour (represented within the \textit{HSB} colour space, using simple computer vision approaches to determine the dominant colour of the object); size (determined by a depth-aware camera); current explorations include shape parametrisation (e.g., following \cite{bechberger2020representing}). A simple data flow diagram (Fig. \ref{dataFlow}) shows the high-level implementation we have used.

Utilisation properties require additional transformations to derive. Sources for these could include task recognition through computer vision techniques (e.g., \cite{feichtenhofer2019slowfast} to determine which tasks are being carried out in the observed video stream (e.g., ‘hammering’ or ‘drilling’); or newer work around the Conceptual Space framework proposing event representation using force and result vectors~\cite{mealier2016construals,gardenfors2018epigenetic}. Currently, utilisation properties are extracted in the training phase using crowdsourced knowledge bases.

\subsubsection{General Knowledge Extraction}
\label{NLP}

We use ConceptNet’s (see \S~\ref{CN}) API\footnote{\url{http://api.conceptnet.io} (accessed on 3 October 2022)} to acquire crowdsourced values for the \textit{UsedFor} property of various manufacturing artefacts (e.g., \textit{drill}). Each response (e.g., \textit{\lq drilling holes in things\rq}) is accompanied by the weight, reflecting the source's reliability. Due to the nature of free-form submissions, messiness of the corresponding response cannot be avoided (e.g., ‘\textit{drilling holes in things}’ and ‘\textit{drill a hole in something}’ are separate entries). To fix that, in the first iteration we group all responses with similar semantics, such as the examples above, and run the softmax function across all weights of semantically different items to acquire the quantity that we use to ground the membership function\footnote{Somewhat counter-intuitively, what is called ‘weight’ in the ConceptNet system is semantically closer to the membership function $\mu$ of the \textit{$\mu$w-model} than the weight $w$. See \S~\ref{muw} for details.} for this utilisation property. Concretely, we get
$$\mu_{A^{Drill}_{UsedFor}}({object\_being\_used\_for\_drilling})=0.9999997$$
which is a reasonable quantity.

Since the described procedure would be quite cumbersome and time-consuming to run for each utilisation property, a more thorough approach to automating utilisation knowledge extraction from the public knowledge bases was developed. It consults ConceptNet and WordNet programmatically, via their respective APIs, alongside the NLTK\footnote{\url{https://www.nltk.org/} (accessed on 4 October 2022)} package, does not require any manual intervention, and proceeds in the following steps:

\begin{enumerate}
    \item Find the appropriate WordNet synset (i.e., a meaning that can be represented by multiple synonymous lexemes) for the given object label (e.g., \textit{'drill'}). The condition is that it denotes an artefact noun. In case there are multiple such synsets, as in the case of \textit{'hammer'} (handheld tool and a gun part), heuristically choose one whose encompassed lexemes have the highest occurrence frequency in the corpus;
    \item Get all synonym lexeme lemmas for that synset using WordNet;
    \item For each lexeme, extract edges from ConceptNet originating from WordNet (higher credibility than crowdsourced data) that have the lexeme as the start node and \textit{'UsedFor'} as the relation. End nodes denote utilisation properties;
    \begin{enumerate}
        \item If no such edge is found (as, e.g., for \textit{forklift}), then consult ConceptNet's crowdsourced knowledge. Of all edges that have the start node and relation as specified, eliminate those whose weight is 1.0 or less (these can be considered unreliable or noise). In each end node of the remaining edges, find a verb using part-of-speech tagger, lemmatise it (e.g., change \textit{'carrying'} to \textit{'carry'}), and add it to the list of utilisations;
        \item If that does not yield a meaningful utilisation property either (as it does not, e.g., for \textit{riveter}), use a heuristic inspired by low inflection of English language: Stem the artifact name (\textit{'riveter'} to \textit{'rivet'}) and check whether it can be used as a verb (\textit{'hammer'}, \textit{'drill'} are both nouns and verbs). If it can, add it to the utilisation list;
    \end{enumerate}
    \item For the extracted verb synsets, extract only those whose meaning falls under WordNet categories \textit{'contact'}, \textit{'change'} or \textit{'motion'} (more can be added where necessary). Synsets characterised as, e.g., \textit{'verb.cognitive'} would be excluded;
    \item Use the retrieved synsets to infer the value for each selected utilisation quality dimension (currently, \textit{'drill'}, \textit{'hammer'}, \textit{'lift'}, \textit{'rivet'}). Do this by comparing these synsets to physical utilisation synsets for each quality dimension, again retrieved from knowledge bases. Where there is a non-empty intersection of synsets, set the value of the corresponding quality dimension to one; where there is not, set it to zero\footnote{Obviously, the proposed heuristics only allow for binary values of the \textit{UsedFor} property. While the current approach served our purpose well, it is a matter of future work to infer continuous values.}.
\end{enumerate}

\begin{center}
  $\ast$~$\ast$~$\ast$
\end{center}

The majority of the described pipeline allows for the implementation of state-of-the-art property detectors. Similarly, the simulation environment could be substituted with real-world cobots with real-world sensors maintaining the same interfaces throughout.

\begin{figure*}[t]
\includegraphics[width=\textwidth]{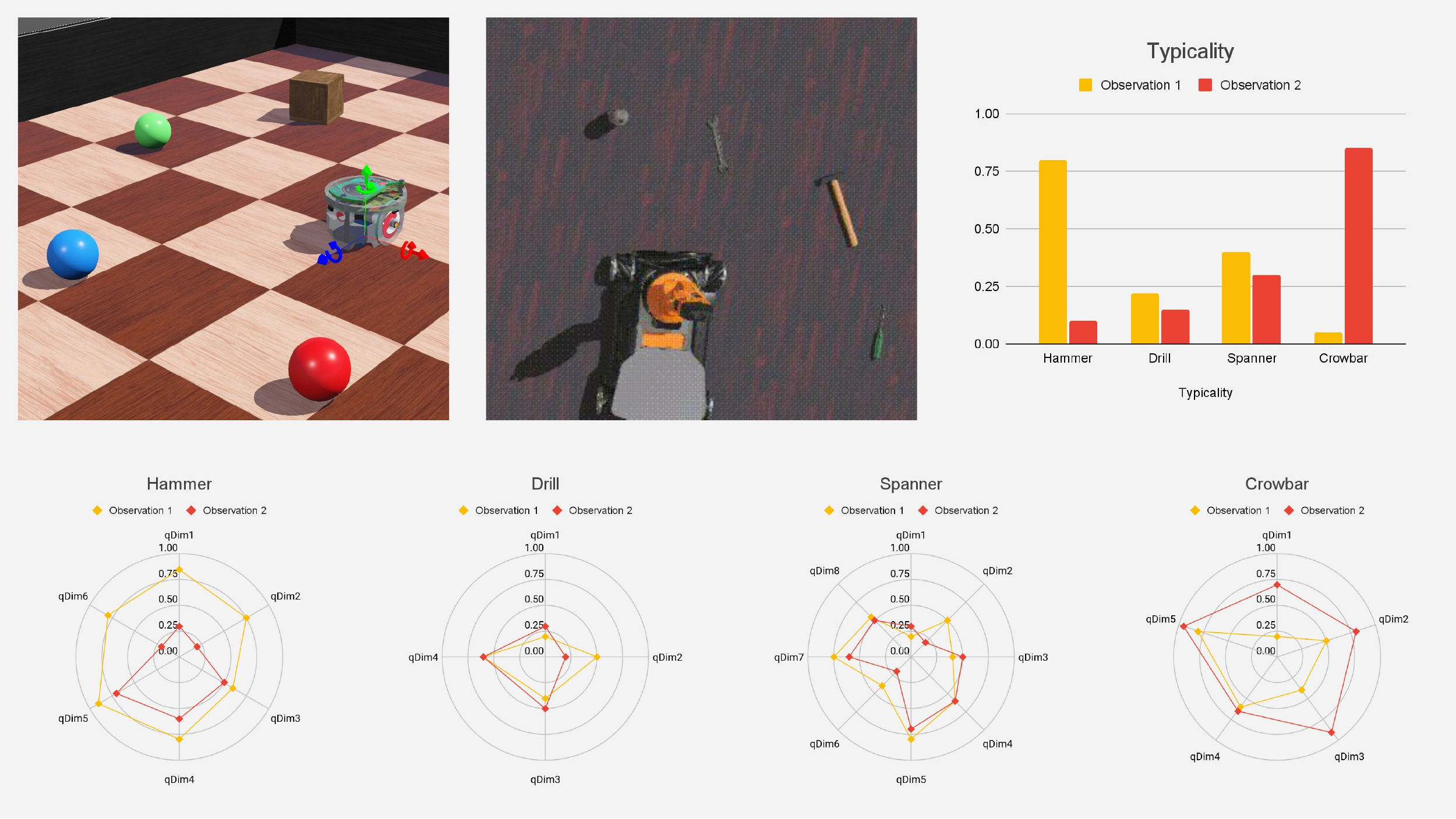}
\caption{Simulated industrial use case environment. We start with the simple case with simulated idealised objects (upper left), such as ‘Red Sphere', ‘Wooden Cube', \textit{etc}. The robot encounters these objects and extracts ground-truth property values available from the simulator (\textit{colour}, \textit{shape}, \textit{composition}), used as the basis for learning the conceptual space. Having validated the data pipeline and knowledge representation modelling, we moved towards the more ecologically valid manufacturing simulated environment (upper middle), where quality dimensions include physical properties like \textit{size}, \textit{texture}, \textit{composition}, and utilisation properties. Generally, every instance's quality dimension values are used as input for membership quantification per various concepts of interest. The spider charts (lower) show example membership values per quality dimensions across four artefact concepts for two example observations. Together with the quality domain weights (currently arbitrated, in the future quantified via empirical hypotheses, see \S~\ref{muw}), these membership values make up the vector representation of the instance, as per Eq.~\ref{eqr}. The bar chart (upper right) represents the typicality of the two example instances across the four observed concepts, calculated via Eq.~\ref{eqR}.}
\label{demo}
\end{figure*}

Quality dimensions’ membership values, along with weights, make up the representativeness vector (Eq.~\ref{eqr}), from which we measure the instance’s typicality across concepts (Eq.~\ref{eqR}) and determine for which one its representativeness is the highest (Eq.~\ref{eqCat}, visualised in Fig.~\ref{demo}). The representativeness score and associated property weights and typicality measures are directly human-comprehensible and constitute our interpretable classification approach, as an (at this point simplified) alternative to traditional computer vision-based object recognition approaches.

\subsection{Situation in the XAI Ecosystem}
\label{widerXAI}
Our knowledge representation-based inherent explainability approach may be situated in a wider XAI framework against roles of system users and previously proposed explanation characteristics. This way it may be compared against other XAI techniques and represent a candidate for the right application context and provided requirements.

Considering development of explainable AI systems teaming with humans in manufacturing environments, the following roles \cite{tomsett2018interpretable} of agents interacting with the AI can be envisioned for such applications:

\begin{itemize}
    \item Creator – Developer of the AI system using property decomposition explainability for sanity check and debugging, as well as guidance in hyperparameter selection (e.g., threshold for disputable classification, see \S~\ref{muw});
    \item Operator and Executor (possibly the same agent) – Manufacturing worker provided with an understandable explanation of the classifier's rationale (see Fig.~\ref{demo}), ready to act upon the output (e.g., take and use the intended fetched tool) or flag the system's incorrect behaviour;
    \item Decision-Subject – In a more advanced use case, a human agent teaming with an embodied AI agent on a physical task (e.g., processing a large component), able to understand the agent's rationale of a tool choice. Obviously, trustworthiness of the AI system is of essence for the Decision-Subject role;
    \item Examiner – Prior to industrialisation and deployment, any AI to be used in applications involving human safety requirements needs to be certified, for which thorough sanity checks of the AI's learnt knowledge and operation rationale are crucial.
\end{itemize}

It should be noted that the expectations from users will likely evolve as interactions increase. For instance, when first working together, it is likely an operator will require more frequent and detailed explanations as they begin to build trust. As this trust develops, it is likely that either less detailed or frequent explanations will be required, and could even cause user frustration (if a system consistently explains a \lq simple\rq{} action undertaken many times per shift). This longitudinal approach is a matter of future work as discussed in \S~\ref{conclusion}.

Our explainability approach can be described across various axes of the explanation characterisation framework, such as one by \cite[Table 1]{hall2019systematic}. Although a rigorous empirical validation of the technique in operation is pending, some characteristics may already be ascertained, e.g.:
\begin{itemize}
    \item Interpretability – High-level interpretability of classification output endowed with visualisation artefacts involving interpretable quality domains and understandable diagrams. As opposed to post-hoc explainability, employed knowledge modelling via decomposition into interpretable quality dimensions is an integral part of the system's design;
    \item Local/global explainability – In the interpretable classification use case (\S~\ref{XC}) it is possible to explore the membership function and weights per quality dimensions and thus validate the categorisation rationale, which adds to global explainability; each instance classification is accompanied by local explainability visual (or textual) artefacts (e.g., Fig.~\ref{demo});
    \item Feature importance quantifiability and visualisation – Learnt knowledge includes empirically validated dimension weights, which provides a cognitively-motivated importance quantification;
    \item Explanation by example – The representational space is partitioned around the prototypes (see \S~\ref{XC}), which may in turn be used to elucidate a seemingly unlikely categorisation outcome;
    \item Interactivity – A simple visual tool has been developed that allows for interactive probing of the classifier by manipulating quality dimension weights and membership functions of the model, as well as property values of an instance;
    \item Counterfactual reasoning – Using the interactive tool (see above) it is possible to understand what the categorisation outcome would have been had the parameters or properties been different.
\end{itemize}

The authors \cite{hall2019systematic} treat the 'generalisability' characteristic as versatility across types of applicable black-box models and as such pertains solely to post-hoc explainability techniques. However, when discussing versatility of utilisation, it is worth noting that the underlying Conceptual Space framework and the proposed typicality model and its parameter learning allow for flexibility of application domain, with an (admittedly strong) assumption of selecting an appropriate set of quality domains and dimensions and their apt modelling. Once these are selected for a given domain, the pertaining parameters are learnt from the combination of exploring in the world (i.e., from observed distributions of quality dimension values) and expert guidance (i.e., providing the labels as input to the concept space construction process).

\subsection{Qualitative Validation}
Empirical validation of interpretability and explainability are still emerging topics in the field of XAI. To try and obtain validation of the work undertaken, we have consulted with subject matter experts and potential users to assess and understand the quality of the of the proposed system.

As discussed in §~\ref{widerXAI}, we identified a number of different user roles, e.g., Creator-Developer, Operator, etc., and presented them with relevant explanations and demonstrations of the system. User feedback was then taken into account to implement improvements. For instance, a number of different visualisations were evaluated for displaying 'typicality' of objects to users (e.g., spider / radar charts, bar charts, doughnut charts, polar charts, etc.), to determine which ones conveyed the most intelligible and useful information.

On top of that, semantics of the interpretable classification output was validated with domain experts, for example, the two-level information presentation. This presentation includes visualisation of the overall calculated representativeness of an instance across categories, which can in turn be 'zoomed-into' to explore typicality across utilised quality dimensions, as in the bar chart and spider charts from Fig.~\ref{demo}, respectively.

Future work will include a rigorous methodological framework for more objective evaluation (on top of subjective, reporting-based approaches) of explainability and interpretability metrics, building on previous domain-relevant and domain-independent work such as \cite{hall2019systematic} and \cite{ZHANG}.

\section{Conclusions and Future Work}
\label{conclusion}

We pursue a knowledge modelling approach towards inherent (vs. post-hoc) explainability of (embodied) AI agents teaming with humans in industrial environments. Heterogeneous knowledge involves physical properties acquirable by equipped sensors and utilisation properties obtained from publicly available crowdsourced and expert-manufactured resources of general knowledge. The properties are consulted by a simple classification model drawing inspiration from Gärdenfors’s Conceptual Space framework. The model is defined by the membership function in the context of fuzzy set theory (quantified frequentistically from experience with environment); and the weight of the property (based on empirically confirmed cognitive semantic hypotheses).

Highly specific industrial environments, such as the aerospace manufacturing one, involve rare and specialist objects, for which there typically do not exist image datasets of sufficient size to train data-hungry high-performant (convolutional) neural networks for object classification. Therefore, our move from computer vision object recognition to property decomposition-based interpretable classification is relevant for such applications characterised by data scarcity, thus necessitating \textit{frugal AI} approaches. Furthermore, decomposing concepts into interpretable property components facilitates model validation and debugging by a developer, scrutinous examination by a certifier, behaviour understandability for a teaming operator, and rationale interpretability for a decision-subject.

Whilst the described use case is a rather limited and illustratory one, it does open a few avenues for future research in knowledge representation and inference modelling of (embodied) AI agents teaming with humans in manufacturing and other industrial and operational environments.

Humans are remarkably successful in learning concepts from scarce data~\cite{hassabis2017neuroscience}, the mechanics and phenomenology of which is of high interest to cognitive science, cognitive neuroscience, computational neuroscience, and artificial intelligence~\cite{kriegeskorte2018cognitive}. One of the challenges is modelling acquisition of core concepts and intuitive theories (e.g., in physics and psychology) as well as generic causal structures~\cite{lake2017building}, all supporting commonsense reasoning. Some promising generative models are based on Bayesian reasoning in the context of hierarchies and structures of hypotheses and associated inductive constraints~\cite{tenenbaum2011grow}. An artificial agent able to acquire and manipulate core concepts arguably makes its behaviour more predictable, its rationales more interpretable, and it itself more trustworthy.

In the simplified examples used we represent quality dimensions via orthogonal basis vectors spanning the conceptual space. However, in reality, property correlations are an important component of concept description \cite{raubal2004formalizing}, stemming from generic, causally originating, core concept structures, and empirically demonstrated by humans’ effortless acquisition of systematic correlations~\cite{billman1996unsupervised,jones2002children,kloos2008s,mcclelland2003parallel}, especially for natural kinds (e.g., many times the colour of fruit predicts its taste and ripeness). It is a matter of further exploration to address property interdependencies, possibly taking into account psychological essentialism, according to which perceptible properties are surface manifestation of entities' true nature~\cite{kornblith1995inductive,galetic2015towards}.

Moreover, for artefacts, which can theoretically take arbitrary values of surface properties, it makes sense to focus on the objects’ purported utilisation capability (\textit{affordances}) stemming from their physical properties, like shape or composition (e.g., a large handheld-sized object with a hard flat metal head is likely to be used to hit nails). Attention-driven visual classification and visualisation techniques may be promising approaches for utilisation inference.

Whilst the current use case deals only with physical objects, newer work around the Conceptual Space framework proposes event representation using force and result vectors~\cite{mealier2016construals,gardenfors2018epigenetic}. This work justifies further exploration as it may prove particularly useful on the shop floor environments. At the same time, one shall need to pay attention to interpretability of pertaining quality domains in this respect.

More recent work in the Conceptual Space theory is focussed on modelling events via force and result vectors~\cite{mealier2016construals,gardenfors2018epigenetic} using the same underlying representational structure as in the described case for (physical) objects. This research is particularly relevant for industrial domains involving human-AI teaming, e.g., for gauging typicality of task execution. A recognised challenge is interpretability of force patterns constituting quality dimensions, which the event-based extension of the Conceptual Space framework is based on.

Cognitive modelling of operators using cognitive architectures~\cite{anderson2004integrated,laird2017standard} is a useful building block for modelling AI cognitive assistants aware of task representations, capable of anomalous behaviour detection, and cognitive load estimation as an input to autonomy engagement. A subsymbolic yet interpretable knowledge representation framework would arguably make a promising step towards modelling the declarative module ((e.g., \cite{lieto2017dual,lieto2017conceptual}), which is yet to be demonstrated in industrial environments involving human-AI teaming.

Reported parallels between Conceptual Spaces' structural principles and empirical findings on conceptual knowledge neural implementation (see §\ref{CS}) invite researchers to explore the appropriateness of such an immanently multidimensional representation for mapping onto what appears to be two-dimensional hexadirectional grid-based encoding system on the neural substrate level\cite{bellmund2018navigating}. Whilst running human studies (e.g., functional magnetic resonance imaging (fMRI) experiments) for hypothesis testing around candidate mapping mechanisms would not be feasible in industrial environments where the authors are affiliated, they shall make every effort to liaise with academic partners who may be better positioned to execute such studies, on top of continually following advances in the area.

Explainability capabilities of an AI system should be able to gauge the human user’s expectations stemming from their experience with the task at hand and the teaming AI agent. Hence a longitudinal approach in context-aware trust and interpretability measurement, possibly involving user’s overt or implicit feedback, and consequent XAI system’s adaptability should be considered in the future.

Finally, it is important to recognise that the path towards a trustworthy AI is multi-faceted and explainability represents one pillar, albeit arguably the one most human user-focussed. The other pillars that are essential for responsible and ethical AI design are robustness and learning assurance, and fairness and non-discrimination\cite{ecai,easa_ml,easa_roadmap}). Obviously, AI trustworthiness should be addressed in an interdisciplinary manner, one difficult without a strong collaboration of academia and industry.

\bibliographystyle{ACM-Reference-Format} 
\bibliography{aamas}

\end{document}